\documentclass{article}

\usepackage{PRIMEarxiv}

\usepackage[utf8]{inputenc} 
\usepackage[T1]{fontenc}    
\usepackage{cite}
\usepackage{hyperref}       
\usepackage{url}            
\usepackage{booktabs}       
\usepackage{amsfonts}       
\usepackage{nicefrac}       
\usepackage{microtype}      
\usepackage{lipsum}
\usepackage{fancyhdr}       
\usepackage{graphicx}       
\graphicspath{{media/}}     
\usepackage{xcolor}

\usepackage{multirow}
\usepackage{array}
\usepackage{caption}
\usepackage{graphicx}
\usepackage{subcaption}
\usepackage{booktabs}
\usepackage{siunitx}
\usepackage{microtype}

\usepackage{afterpage}

\usepackage[dvipsnames,svgnames]{xcolor}

\definecolor{darkgreen}{named}{DarkGreen}
\definecolor{darkorange}{named}{orange}
\definecolor{darkred}{named}{red}

\newcommand{\textsentgreen}[1]{\textcolor{darkgreen}{#1}}

\newcommand{\textsentred}[1]{\textcolor{darkred}{#1}}

\usepackage[normalem]{ulem}

\usepackage{amssymb}

\newcommand{\NoteSent}[2]{\texttt{\textbf{#1:}} #2}
\newcommand{\NoteSentGreen}[2]{\texttt{\textbf{#1:}} \textsentgreen{#2}}

\newcommand{\CitationGreen}[1]{\textsentgreen{\texttt{\textbf{[#1]}}}}

\newcommand{\CitationRed}[1]{\textsentred{\texttt{\textbf{[#1]}}}}

\usepackage{silence}
\usepackage{tocloft}

\cftpagenumbersoff{figure}

\usepackage{placeins}

\usepackage{pdfpages}

\title{Automated Evaluation can Distinguish the Good and Bad AI Responses to Patient Questions about Hospitalization}

\author{Sarvesh Soni and Dina Demner-Fushman \\
  Division of Intramural Research \\
  National Library of Medicine, National Institutes of Health \\
  Bethesda, MD, USA\\
  \texttt{sarvesh.soni@nih.gov}, \texttt{ddemner@mail.nih.gov}
}

\begin{document}
\maketitle

\begin{abstract}
Automated approaches to answer patient-posed health questions are rising, but selecting among systems requires reliable evaluation.
The current gold standard for evaluating the free-text artificial intelligence (AI) responses--human expert review--is labor-intensive and slow, limiting scalability.
Automated metrics are promising yet variably aligned with human judgments and often context-dependent.
To address the feasibility of automating the evaluation of AI responses to hospitalization-related questions posed by patients, we conducted a large systematic study of evaluation approaches.
Across $100$ patient cases, we collected responses from $28$ AI systems ($2800$ total) and assessed them along three dimensions: whether a system response (1) answers the question, (2) appropriately uses clinical note evidence, and (3) uses general medical knowledge.
Using clinician-authored reference answers to anchor metrics, automated rankings closely matched human ratings.
Our findings suggest that carefully designed automated evaluation can scale comparative assessment of AI systems and support patient-clinician communication.
\end{abstract}

\section{Introduction}

Patients are increasingly turning to generative artificial intelligence (AI) to seek medical information and guidance \cite{shahsavar2023UserIntentionsUse,busch2025CurrentApplicationsChallenges,soni2025OverviewArchEHRQA2025}.
Yet patients are not well-positioned to independently assess the truthfulness, reliability, and clinical appropriateness of model outputs \cite{osnat2025PatientPerspectivesArtificial}, underscoring the need for robust mechanisms to evaluate healthcare AI systems \cite{bedi2025TestingEvaluationHealth}.
For their ease in automated scoring, repeatability, and cross-system comparison, many evaluations rely on multiple-choice question (MCQ) benchmarks (e.g., USMLE \cite{kung2023PerformanceChatGPTUSMLE}).
However, emerging evidence indicates that MCQ-based assessments are vulnerable to format-specific artifacts and superficial pattern matching, potentially inflating apparent performance and obscuring deficits in clinical reasoning \cite{balepur2024ArtifactsAbductionHow,griot2025PatternRecognitionMedical}.
Evaluation using long-form generation by the AI models is shown to be more reliable than MCQs \cite{li2024CanMultiplechoiceQuestions}.
In the medical domain, however, long-form evaluation has remained small-scale because high-quality assessment of free-text AI responses requires deep subject-matter expertise and careful rubric design \cite{ayers2023ComparingPhysicianArtificial,bernstein2023ComparisonOphthalmologistLarge}.
Given the increasing number of models \cite{naveed2025ComprehensiveOverviewLarge} and the growing number of clinical applications built on top of them \cite{liu2025ApplicationLargeLanguage}, it is impractical to scale human annotation at the pace of model development.
There is a pressing need for evaluation strategies that balance human oversight with scalable automation.

Automated evaluation of free-text responses, especially for question answering (QA), remains challenging \cite{farea2025EvaluationQuestionAnswering}.
Prior work has examined correlations between automated metrics (including n-gram or embedding similarity and large language model [LLM]-based judges) and human judgments \cite{sai2022SurveyEvaluationMetrics,moramarco2022HumanEvaluationCorrelation}.
Yet metric choice is highly context-dependent \cite{sai2022SurveyEvaluationMetrics}, motivating targeted studies to investigate the suitability of automated evaluation in patient-centered QA where safety and calibration matter.
Moreover, few studies explicitly test whether automated metrics can recover human rank-order preferences among system outputs, e.g., via pairwise human evaluations \cite{chen2024MLLMasaJudgeAssessingMultimodal}.
Complementary approaches have explored the use of information nuggets to automatically evaluate system responses \cite{marton2006NuggeteerAutomaticNuggetBased}, but constructing a comprehensive set of nuggets is labor-intensive and rarely exhaustive \cite{bartels2025CanLargeLanguage}.

\begin{table}[h!]
\centering
\caption{Description of human evaluation dimensions.}
\label{tab:evaluation-dimensions}
\renewcommand{\arraystretch}{1.25}
\begin{tabular}{l p{2cm} p{12cm}}
\hline
\multicolumn{3}{l}{\textbf{Criterion 1: Does the output answer the question?} \textit{(answers-question)}} \\ \hline
& Yes       & Output includes a full, direct, and explicit answer to the provided question. \\
& Partially & Output includes only a partial answer or implies, but does not directly state the answer. \\
& No        & Output does not answer the question directly or indirectly. \\ \hline

\multicolumn{3}{l}{\textbf{Criterion 2: Does the output use the EHR evidence? And how?} \textit{(uses-evidence)}} \\ \hline
& Yes     & At least one assertion is supported by or inferred from the evidence. \\
& No      & No assertions in the output use the evidence. \\
& Refutes & At least one assertion is refuted by the evidence. \\ \hline

\multicolumn{3}{l}{\textbf{Criterion 3: Does the output use any general knowledge?} \textit{(uses-knowledge)}} \\ \hline
& Yes        & Uses general knowledge that is in line with or neutral to the question. \\
& No         & Does not include any general knowledge. \\
& Conflicting & Provides contradictory or incorrect knowledge (including hallucinations). \\ \hline
\end{tabular}
\end{table}

To address these gaps, we conduct a large-scale, multi-dimensional human evaluation of AI responses to patient questions about their hospitalization.
We systematically benchmark multiple \emph{answer-generation systems}, i.e., end-to-end AI pipelines that take the same patient question and clinical note excerpt as input and produce free-text answers with sentence-level citations to the provided evidence.
We evaluate each answer along three clinically salient dimensions—\textit{answers-question}, \textit{uses-evidence}, and \textit{uses-knowledge}—and derive system-level rankings by aggregating three independent annotations per answer using \textit{Pyramid} (summing numeric label scores across annotators per case) and \textit{MACE} (Multi-Annotator Competence Estimation; an unsupervised probabilistic label-aggregation model that jointly infers annotator reliability and a latent true label for each response).
We then compare system rankings derived from human evaluation with those produced by automated metrics.
We consider two types of automated metrics: \textit{Factuality} metrics assess citation alignment with gold sentence-level relevance labels (reported as citation precision/recall/F1, with \textit{strict} and \textit{lenient} variants), while \textit{Relevance} metrics compare generated answer text to a reference using lexical or semantic similarity.
We compute \textit{Relevance} either against clinician-authored reference answers (denoted by \textit{``(human)''}) or against a proxy reference formed by concatenating the note sentences labeled as \textit{essential} with the input question (denoted by \textit{``(note)''}).
Taken together, our study provides empirical evidence and practical guidance for when and how automated metrics can reduce reliance on costly human evaluation, thereby informing evaluation strategies for patient-centered medical AI.

Our contributions are threefold.
First, we present a large-scale human evaluation of $2800$ evidence-grounded answers from $28$ systems across $100$ real-world patient cases, producing $25{,}200$ judgments over three clinically meaningful dimensions.
Second, we establish a system-level benchmarking framework that compares automated metrics to human judgments via rank-order agreement and assesses ranking robustness across aggregation methods (Pyramid vs.\ MACE).
Third, we provide empirical guidance on which metric families and reference choices best recover human \emph{system-level} rankings across dimensions, highlighting the importance of evaluation dimension and reference type.

\section{Results}

For each of the $100$ patient cases, we collected responses from $28$ different AI systems, yielding a total of $2800$ AI responses \cite{soni2025OverviewArchEHRQA2025}.
Each AI response was independently evaluated by three medical student annotators on three dimensions (Table \ref{tab:evaluation-dimensions}), resulting in $8400$ judgments per dimension.
Table \ref{tab:label-distributions} summarizes the distribution of annotation labels across dimensions.
On the \textit{answers-question} dimension, most system responses were labeled \textit{yes} ($45.5\%$) or \textit{partially} ($31.4\%$).
Severe error labels were relatively rare: on \textit{uses‑evidence}, $7.3\%$ of judgments were labeled \textit{refutes}, and on \textit{uses‑knowledge}, $3.8\%$ were labeled \textit{conflicting}.
A majority vote (at least two judgments out of three) of annotators yielded a \textit{yes} decision for \textit{answers‑question} in $45.9\%$ of cases and for \textit{uses‑evidence} in $88.3\%$ of cases.

\begin{table}[h!]
  \centering
  \caption{Label distributions by dimension and scope. \textit{Overall:} label counts out of $8400$ total original annotations ($2800$ responses × $3$ annotators). \textit{Majority:} system response counts where a majority of annotators agreed out of $2800$ total responses; percentages may not sum to $100\%$ due to ties. \textit{MACE:} label counts after automated reconciliation using MACE (Multi-Annotator Competence Estimation \cite{hovy2013LearningWhomTrust}) out of $2800$ total reconciled labels. All values are in \textit{count (percent)} format.}
  \renewcommand{\arraystretch}{1.25}
  \label{tab:label-distributions}
  \resizebox{\textwidth}{!}{
  \begin{tabular}{l ccc ccc ccc}
    \toprule
    & \multicolumn{3}{c}{\textit{answers-question}} & \multicolumn{3}{c}{\textit{uses-evidence}} & \multicolumn{3}{c}{\textit{uses-knowledge}} \\
    \cmidrule(lr){2-4} \cmidrule(lr){5-7} \cmidrule(lr){8-10}
    Scope & \textit{yes} & \textit{partial} & \textit{no} & \textit{yes} & \textit{no} & \textit{refute} & \textit{yes} & \textit{no} & \textit{conflict} \\
    \midrule
    Overall & 3824 (45.5) & 2634 (31.4) & 1942 (23.1) & 6941 (82.6) & 844 (10.0) & 615 (7.3) & 2291 (27.3) & 5790 (68.9) & 319 (3.8) \\
    Majority & 1286 (45.9) & 767 (27.4) & 580 (20.7) & 2472 (88.3) & 159 (5.7) & 101 (3.6) & 583 (20.8) & 2087 (74.5) & 40 (1.4) \\
    MACE & 1278 (45.6) & 759 (27.1) & 763 (27.3) & 2038 (72.8) & 369 (13.2) & 393 (14.0) & 950 (33.9) & 1626 (58.1) & 224 (8.0) \\
    \bottomrule
  \end{tabular}
  }
\end{table}

\begin{table}[h!]
  \centering
  \caption{Annotation agreement summary across dimensions. \textit{Three Labels:} agreement using the original three labels for each dimension. \textit{Binary Labels:} agreement after dichotomizing the labels into \textit{yes} (merging \textit{partially} into \textit{yes} for \textit{answers-question}) or \textit{no} (merging \textit{refutes} and \textit{conflicting} into \textit{no} for \textit{uses-evidence} and \textit{uses-knowledge}). Values show unanimous agreement rate, mean pairwise agreement, and Fleiss' $\kappa$.}
  \renewcommand{\arraystretch}{1.25}
  \label{tab:agreement}
  \begin{tabular}{l ccc ccc}
    \toprule
     & \multicolumn{3}{c}{\textit{Three Labels}} & \multicolumn{3}{c}{\textit{Binary Labels}} \\
    \cmidrule(lr){2-4} \cmidrule(lr){5-7}
    Dimension & Unanimous & Pairwise & $\kappa$ & Unanimous & Pairwise & $\kappa$ \\
    \midrule
    \textit{answers-question} & 40.2\% & 58.1\% & 0.347 & 75.7\% & 83.8\% & 0.545 \\
    \textit{uses-evidence} & 66.0\% & 76.5\% & 0.222 & 67.1\% & 78.1\% & 0.236 \\
    \textit{uses-knowledge} & 45.0\% & 62.2\% & 0.159 & 49.2\% & 66.1\% & 0.147 \\
    \bottomrule
  \end{tabular}
\end{table}


Table \ref{tab:agreement} reports inter‑annotator agreement.
Exact three‑way agreement occurred $40.2\%$, $66.0\%$, and $45.0\%$ of the time for \textit{answers‑question}, \textit{uses‑evidence}, and \textit{uses‑knowledge}, respectively.
After binarizing labels to \textit{yes} vs. \textit{no}, three‑way agreement increased to $75.7\%$, $67.1\%$, and $49.2\%$, respectively.
Mean pairwise agreement was $58.1\%$, $76.5\%$, and $62.2\%$ for the three dimensions and similarly rose after binarization.
Fleiss' $\kappa$ values were $0.347$, $0.222$, and $0.159$ for \textit{answers‑question}, \textit{uses‑evidence}, and \textit{uses‑knowledge}, respectively, indicating moderate chance-corrected agreement for \textit{answers-question} and lower agreement for the other two dimensions, consistent with the subjective nature of open-ended clinical text judgments.

\begin{figure}[t!]
    \begin{center}
        \includegraphics[width=\textwidth]{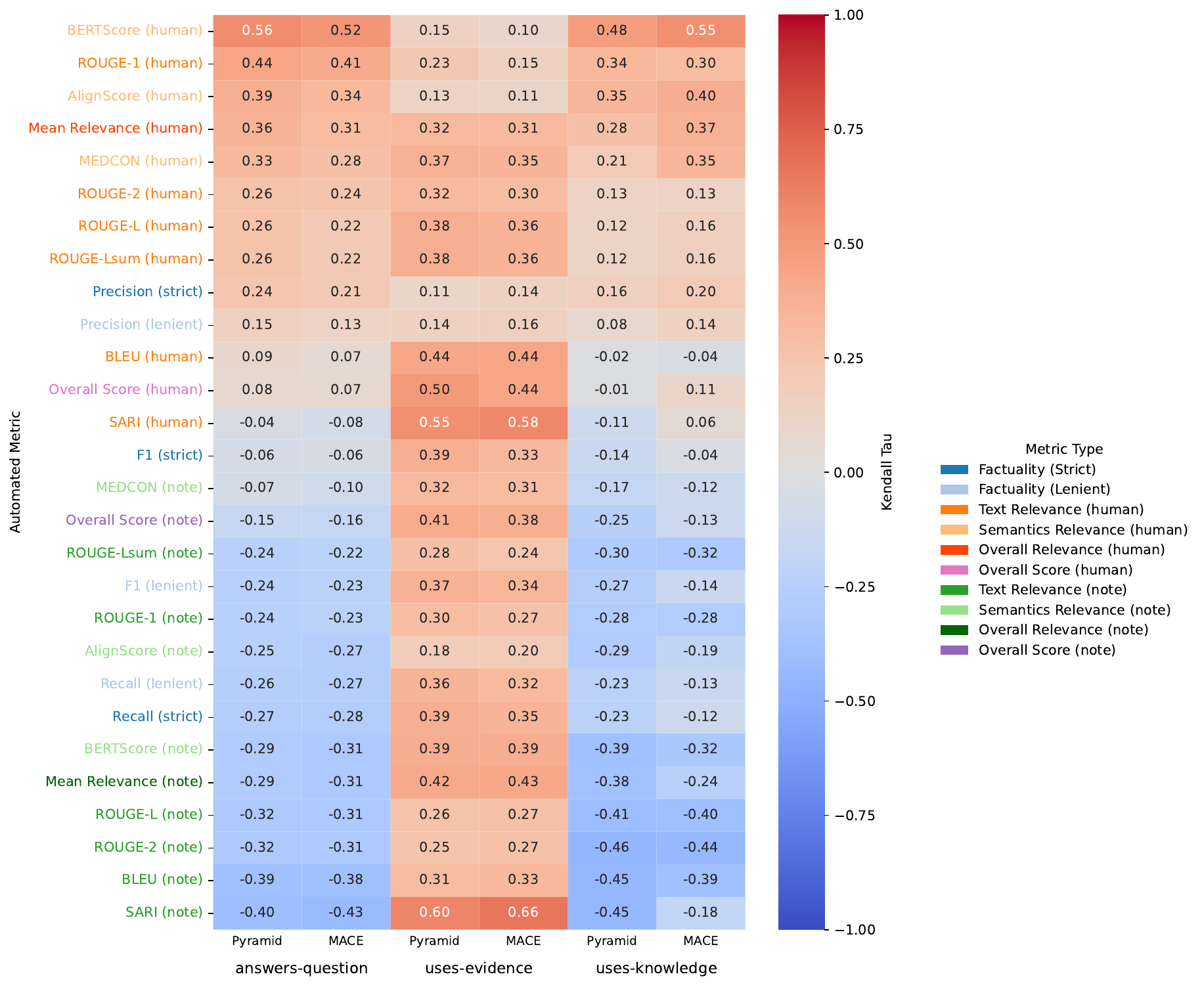} 
        \caption{Kendall's $\tau$ correlations between system rankings induced by automated metrics (rows) and human‑judgment rankings under two rank‑aggregation schemes (\textit{Pyramid} and \textit{MACE}) across the three evaluation dimensions. Cells encode $\tau$ (-1 to +1; blue $\rightarrow$ red). Rows are grouped and color-coded by metric type (legend) and are ordered by $\tau$ for \textit{answers‑question} under the \textit{Pyramid} scheme. Metrics labeled \textit{``(human)''} are computed against the clinician-authored reference answer, while metrics labeled \textit{``(note)''} are computed against a proxy reference formed by concatenating the note sentences annotated as \textit{essential} with the input question.}
        \label{fig:rank-corr}
    \end{center}
\end{figure}

Figure \ref{fig:rank-corr} shows Kendall's $\tau$ correlations between system rankings induced by automated metrics and by human judgments.
For \textit{answers‑question}, \textit{Relevance} metrics that compare system outputs to clinician‑authored reference answers yielded the strongest correspondence: \textit{BERTScore (human)} ($\tau$=$0.56$ with \textit{Pyramid}; $0.52$ with \textit{MACE}), \textit{ROUGE‑1 (human)} ($0.44$; $0.41$), and \textit{AlignScore (human)} ($0.39$; $0.34$).
In contrast, \textit{Relevance} metrics computed against the note-based proxy reference were weakly or negatively associated with human rankings, e.g., \textit{SARI (note)} ($\tau$=$-0.40$; $-0.43$) and \textit{BLEU (note)} ($-0.39$; $-0.38$), suggesting that note-based similarity is a poor proxy for \textit{answers‑question}.
Interestingly, this highlights strong reference dependence: the association for the same metric can even flip sign, e.g., \textit{BERTScore (human)} shows a positive association while \textit{BERTScore (note)} shows a negative association.
In contrast, \textit{Relevance} metrics computed against the note-based proxy reference were weakly or negatively associated with human rankings (e.g., \textit{SARI (note)} ($\tau$=$-0.40$; $-0.43$) and \textit{BLEU (note)} ($-0.39$; $-0.38$)), and the association can even flip sign with reference choice (e.g., \textit{BERTScore (human)} positive vs.\ \textit{BERTScore (note)} negative), suggesting that note-based similarity is a poor proxy for \textit{answers-question}.

The pattern reverses for \textit{uses‑evidence}, where text‑overlap metrics against input note correlate most strongly with human rankings: \textit{SARI (note)} ($0.55$–$0.66$), \textit{BLEU (note)} ($0.31$–$0.44$), and \textit{ROUGE‑L (note)} ($0.26$–$0.38$).
\textit{Factuality} metrics based on citation alignment showed moderate correlation: citation \textit{F1} ($\tau$=$0.33$–$0.39$) and \textit{Recall} ($0.32$–$0.39$) correlated more strongly than \textit{Precision} ($0.11$–$0.16$).
For \textit{uses‑knowledge}, \textit{Relevance} metrics computed against clinician reference answers again performed best, with \textit{BERTScore (human)} ($0.48$; $0.55$) and \textit{AlignScore (human)} ($0.35$; $0.40$).
These trends are broadly consistent across the two rank‑aggregation methods (\textit{Pyramid} vs. \textit{MACE}) with the sign of association usually preserved, though a few metrics show larger method‑dependent differences (most notably \textit{SARI (note)} for \textit{uses‑knowledge} with $\tau$=-$0.45$ vs. -$0.18$).

\begin{figure}[t!]
    \begin{center}
        \includegraphics[width=\textwidth]{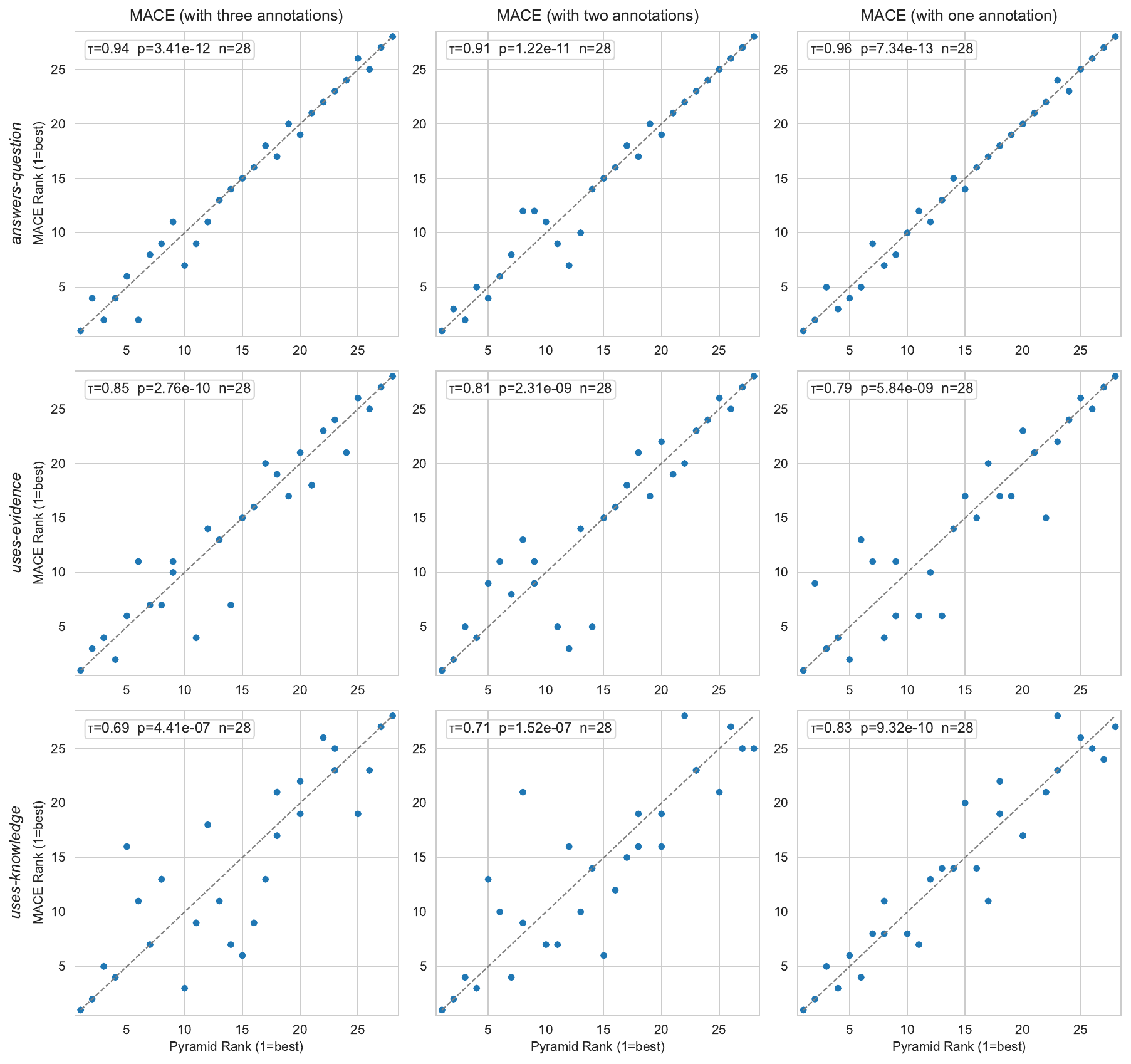} 
        \caption{Agreement between Pyramid and MACE rankings. Each panel (rows: \textit{answers‑question}, \textit{uses‑evidence}, \textit{uses‑knowledge}; columns: \textit{MACE with three annotations}, \textit{MACE with two annotations}, \textit{MACE with one annotation}) plots MACE rank (y; 1=best) against Pyramid rank (x; 1=best) for n=28 systems; the dashed line marks perfect agreement. ``MACE with three annotations'' uses all three annotations per answer; ``MACE with $k$ annotations'' is obtained by subsampling $k$ of the three available annotations per answer to simulate constrained annotation budgets. Kendall's $\tau$ is annotated in each panel.}
        \label{fig:pyramid-mace-corr}
    \end{center}
\end{figure}

Figure \ref{fig:pyramid-mace-corr} demonstrates agreement between the two ranking schemes.
Across the three dimensions, \textit{Pyramid}‑based and \textit{MACE}‑based system rankings were strongly concordant (Kendall's $\tau$ $0.69$–$0.96$, n=$28$, $p \leq 4.41 \times 10^{-7}$).
For \textit{answers‑question}, correlations were particularly high between \textit{Pyramid} and \textit{MACE} using all three annotations ($\tau$=$0.94$), as well as when simulating smaller annotation budgets by subsampling to \textit{MACE using 2 annotations} ($0.91$) and \textit{MACE using 1 annotation} ($0.96$).
The weakest, yet still significant, concordance occurred for \textit{uses‑knowledge} under \textit{MACE} ($\tau$=$0.69$).
Scatterplots cluster along the $45^\circ$ line, indicating near‑monotonic alignment of ranks.
Notably, \textit{MACE} with only one or two annotations closely tracked \textit{Pyramid}, suggesting that robust rank‑order recovery is possible under constrained annotation budgets.
Differences in $\tau$ across the subsampled settings are small and can arise from subsampling variability; the main observation is that rank order remains strongly preserved.

\begin{figure}[t!]
    \begin{center}
        \includegraphics[width=\textwidth]{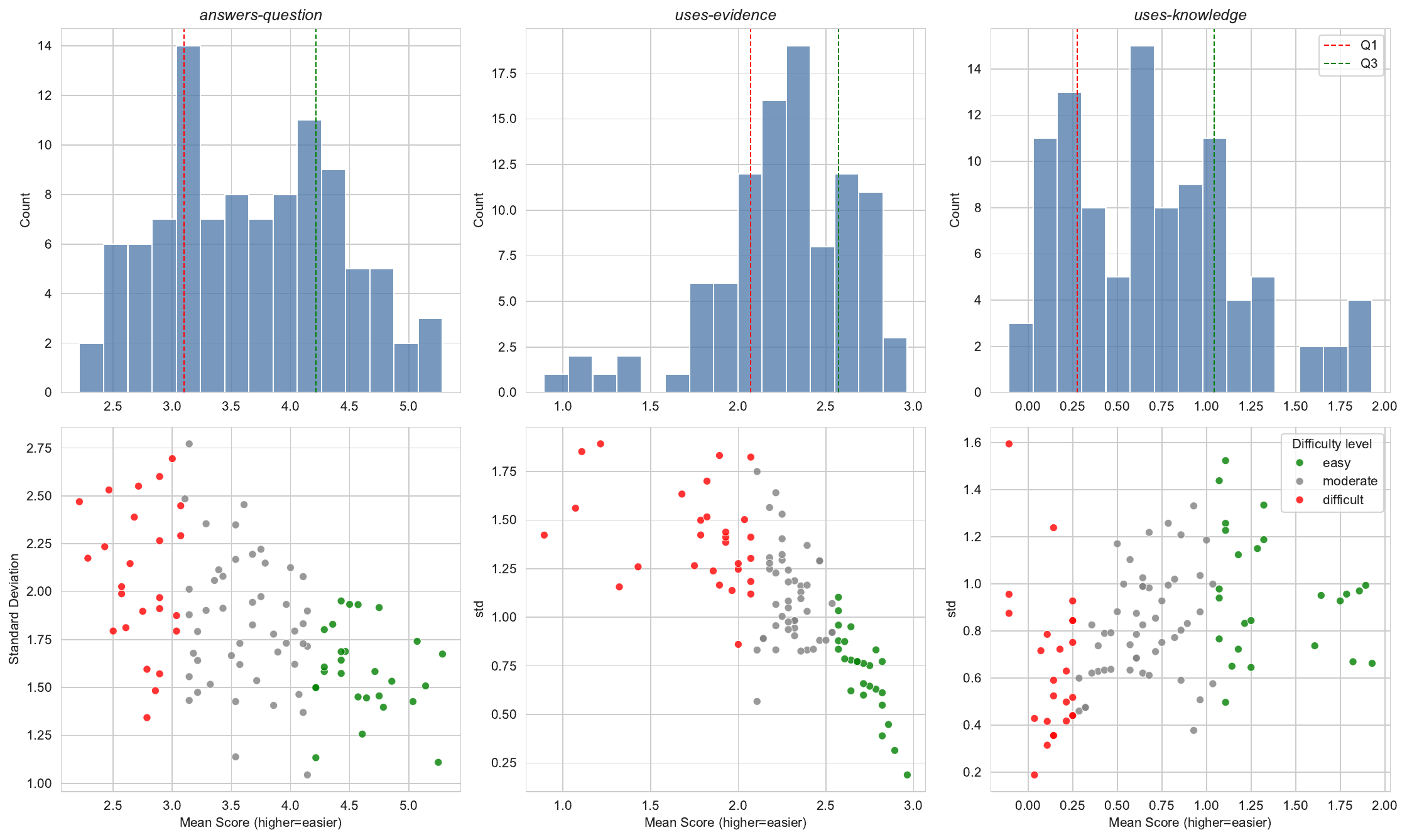} 
        \caption{Case‑level performance across systems using the Pyramid scheme. Top row: histograms of across‑system mean scores for each case on the three dimensions (\textit{answers‑question}, \textit{uses‑evidence}, \textit{uses‑knowledge}); vertical dashed lines mark the first (Q1) and third (Q3) quartiles. Bottom row: per‑case mean score (higher = easier) versus across‑system standard deviation, with points colored by difficulty level (scatter plot legend). Cases are labeled as difficult if their mean falls at or below Q1, easy if it falls at or above Q3, and moderate otherwise.}
        \label{fig:case-analysis}
    \end{center}
\end{figure}

Figure \ref{fig:case-analysis} presents case‑level performance profiles across systems for all three evaluation dimensions.
Easier cases (means above the third quartile) exhibited lower across‑system dispersion, whereas harder cases (means below the first quartile) showed higher dispersion.
This inverse association between case difficulty and inter‑system variability implies that when questions are easier, models not only score higher but also agree more closely.
Dimension‑specific ranges also differed (mid‑range mean for \textit{answers‑question}, higher for \textit{uses‑evidence}, lower for \textit{uses‑knowledge}), delineating distinct challenge profiles across dimensions.

\section{Discussion}

We systematically evaluated the correlations of system rankings induced by automated \textit{Factuality} and \textit{Relevance} metrics versus human judgments across three distinct evaluation dimensions for the task of grounded QA using clinical notes.
We collected human judgments for the free-text responses from $28$ AI systems across $100$ patient cases ($2800$ total), yielding $8400$ annotations per dimension ($25{,}000$ total) across three evaluation dimensions, namely, \textit{answers-question}, \textit{uses-evidence}, and \textit{uses-knowledge}.
Using our systematic analyses, we presented the performance of the evaluated systems, identified the context- and input-dependent use of different automated metrics for the EHR QA task, established the usefulness of the employed automatic reconciliation method, and categorized the case-level performance across systems.

We found that having a ground truth expert-curated answer to the test questions provides a strong signal for ranking the automated systems for the task of automatically responding to patient questions with free-text answers.
Specifically, among \textit{Relevance} metrics, semantic similarity (e.g., BERTScore) computed against clinician-authored reference answers is a good substitute for manually ranking each system response for the \textit{answers-question} and \textit{uses-knowledge} dimensions, while the text-overlap metrics (e.g., SARI) against the input evidence better explain the \textit{uses-evidence} dimension.

BERTScore between system-generated and ground truth text is shown to correlate better with human judgments \cite{zhang2019BERTScoreEvaluatingText}.
Notably, the use of different ground truth texts (clinician-answers vs. note sentences) substantially altered correlations with human judgments, particularly for the \textit{answers-question} dimension.
This aligns with prior work showing that BERTScore is sensitive to the choice of reference \cite{hanna2021FineGrainedAnalysisBERTScore,nguyen2024ReferencebasedMetricsDisprove}.
Moreover, its correlations varied across evaluation dimensions, indicating that a single similarity score may not be equally informative for all aspects of QA.
Accordingly, automated metrics are context- and reference-dependent; human evaluation remains essential to establish or refresh the reference standard and to validate metrics in the target setting.
Nevertheless, our results suggest that investing substantially less effort to obtain clinician-authored reference answers (rather than judging all system responses) can still provide a strong signal for ranking systems on the \textit{answers-question} dimension.

To assess alignment between automated metrics and human judgments, we compared the rankings induced by each metric with those from human evaluation rather than correlating raw scores.
This choice targets ordinal agreement (the quantity most relevant for comparative benchmarking) and reduces sensitivity to differences in scale, dynamic range, and nonlinear relationships.
Consistent with prior work, we observed stronger correspondence at the system level than at the level of individual human scores, reinforcing that automatic metrics are more reliable for system‑level comparisons than for item‑level judgments \cite{novikova2017WhyWeNeed}.
Given the pace at which new systems appear, the practical objective is to rank contenders accurately with minimal additional annotation.
Accordingly, system‑level rank correlations provide a concise indicator of metric utility.

Notably, text‑overlap \textit{Relevance} metrics (e.g., SARI) aligned more closely with the \textit{uses-evidence} dimension than \textit{Factuality} metrics based on citation F1 scores.
A likely explanation is architectural: nearly all evaluated systems implement an explicit evidence‑selection stage (identifying and attaching citations) before producing the final answer \cite{soni2025OverviewArchEHRQA2025}.
Consequently, citation‑F1 predominantly reflects the fidelity of this retrieval/citation-selection module rather than the extent to which the generated narrative actually leverages the cited material.
By contrast, text‑oriented metrics are more sensitive to whether evidence is integrated and transformed within the response itself.
These findings caution against equating high citation‑F1 with effective evidence use and support evaluation protocols that decouple retrieval from generation, reporting complementary measures for both selection fidelity and evidence integration in text.

We view human and automated evaluation as complementary, and our findings do not suggest replacing human evaluation in safety-critical clinical settings.
Rather, human judgments are essential for establishing high-quality references (e.g., clinician-authored answers or relevance labels) and for conducting error analysis of clinical errors and omissions not captured by similarity metrics.
Once such references are available, automated metrics can serve as a scalable screening and benchmarking layer for rapid system comparison and regression monitoring, reserving expert review for ambiguous or high-risk outputs.

\begin{table}[t]
    \footnotesize
    \centering
    \setlength{\tabcolsep}{4pt}
    \renewcommand{\arraystretch}{1.1}
    \caption{Example annotated patient case with sample system responses. In the note excerpt, sentences \texttt{[5,6]} are \textsentgreen{\texttt{essential}} and the rest are \textsentred{\texttt{not-relevant}}.}
    \begin{tabular}{p{0.08\linewidth} p{0.88\linewidth}}
        \toprule
        \multicolumn{2}{c}{\textbf{Example Case}} \\
        \midrule
        \textbf{Patient\newline Question} & Hi Dr. My mom is 88 years old, no heart problems. She hit her head and was hospitalized. No intubation was done in ICU. Her blood count and pressure were normal, she could eat and was breathing on her own. My question is about the antibiotics she was given during her hospital stay. Were they necessary? Thank you, \\
        \midrule
        \textbf{Clinician\newline Question} & Why was she given antibiotics during her hospital stay? \\
        \midrule
        \textbf{Clinical \newline Note \newline Excerpt} &
        \NoteSent{1}{Discharge Instructions:}        
        \NoteSent{2}{At first, you were in the ICU, but when you improved, you were transferred to the regular neurology floor for care.}
        \NoteSent{3}{After someone hits their head, we watch them very closely for signs of worsened swelling or bleeding in the brain.}
        ~\NoteSent{4}{All of your scans after the fall were relatively stable, and you made great progress with your level of alertness, memory, and physical activity in the days following the accident.}

        \par\vspace{0.5\baselineskip}
        
        \NoteSentGreen{5}{On 2174-2-11, it was noted that you were not holding your oxygen levels as high as we would like so a chest x-ray was done, and indicated that you were developing a pneumonia.}
        ~\NoteSentGreen{6}{We started IV antibiotics for this right away, and you improved.}

        \par\vspace{0.5\baselineskip}
        
        \NoteSent{7}{Physical and occupational therapy saw you during this hospital stay, and recommended a short stay in rehab before you return to your home.}
        ~\NoteSent{8}{At rehab, you will work to become stronger and safer on your own.} \\
        \midrule
        \textbf{Clinician \newline Answer} &
        The patient was given antibiotics during her hospital stay because of a lung infection \CitationGreen{5}. The patient was noted to have low levels of oxygen and a chest x-ray was ordered \CitationGreen{5}. it indicated the patient was developing pneumonia \CitationGreen{6}. IV antibiotics were initiated to treat the pneumonia \CitationGreen{6}. The patient showed improvement right away after starting the antibiotics \CitationGreen{6}. Therefore, the antibiotics were necessary during her hospital stay. \\
        \midrule
        \multicolumn{2}{c}{\textbf{Sample Response \#1} (rated worst on \textit{answers-question})} \\
        \midrule
        \textbf{System \newline Response} &
        She was given antibiotics during her hospital stay because there is no information available regarding the necessity of these medications. \CitationRed{1} However, discharge instructions indicate that some form of treatment plan may have been implemented. \CitationRed{1} \\
        \textbf{Comments} &
        Does not directly or indirectly answer the question. \\
        \midrule
        \multicolumn{2}{c}{\textbf{Sample Response \#2} (rated worst on \textit{uses-evidence} and \textit{uses-knowledge})} \\
        \midrule
        \textbf{System \newline Response} &
        The antibiotics were given promptly as a precautionary measure to prevent potential infections following the head injury, which may or may not have resulted in a mild traumatic brain injury. \CitationRed{3,7} This strategy aimed to mitigate the risk of secondary infections and promote overall recovery, given the close monitoring and improvement observed in the ICU and subsequent transfer to the neurology floor. \textsentred{\texttt{\textbf{[2,}}}\textsentgreen{\texttt{\textbf{6]}}} \\
        \textbf{Comments} &
        Refutes the evidence and general knowledge about the use of antibiotics for this patient. \\
        \midrule
        \multicolumn{2}{c}{\textbf{Sample Response \#3} (rated best on all three dimensions)} \\
        \midrule
        \textbf{System \newline Response} &
        The patient was given antibiotics during her hospital stay due to the development of pneumonia, as indicated by the chest x-ray on 2174-2-11. \CitationGreen{5} This pneumonia was a concern for the healthcare team, as it could have potentially led to further complications if left untreated. \CitationGreen{5} The antibiotics were started immediately to address the pneumonia and improve the patient's condition. \CitationGreen{6} The patient's response to the antibiotics was positive, as she improved following treatment. \CitationGreen{6} \\
        \textbf{Comments} &
        Answers the question directly, uses the provided clinical evidence, and uses general medical knowledge. \\
        \bottomrule
    \end{tabular}
    \label{tab:example}
\end{table}


Table \ref{tab:example} presents sample system responses for a patient case from the dataset.
On the \textit{answers-question} dimension, the primary cause of low ratings was the system's tendency to rephrase input evidence (either in support of or in opposition to it) without directly addressing the patient's inquiry.
On the \textit{uses-evidence} and \textit{uses-knowledge} dimensions, the worst-rated system responses made claims inconsistent with the input questions or the clinical note and often employed conflicting world knowledge.

Given the dimension-specific differences in Table~\ref{tab:agreement}, we report both chance-corrected agreement (Fleiss' $\kappa$) and complementary raw/binarized agreement to characterize reliability across dimensions.
These agreement patterns are expected for open-ended clinical text judgments that involve partial credit, implicit reasoning, and different interpretations.
More broadly, disagreement in clinical interpretation is well documented even among specialists \cite{novack2006DisagreementInterpretationChest}.
Related open-ended tasks, such as Medical Subject Headings (MeSH) indexing (assignment of MeSH terms to scientific articles) and evaluation of model-generated clinical notes, similarly report modest to low inter-rater agreement \cite{funk1983IndexingConsistencyMEDLINE,fernandez-llimos2025ConsistencyMedicalSubject,benabacha2023InvestigationEvaluationMethods,benabacha2023EmpiricalStudyClinical,fleming2024MedAlignClinicianGeneratedDataset,moramarco2022HumanEvaluationCorrelation}.
Thus, the agreement levels observed here are within the range reported for comparable biomedical text evaluation settings.
Furthermore, chance-corrected coefficients such as $\kappa$ can also be deflated under strong label imbalance (the ``high agreement/low-$\kappa$'' paradox) \cite{feinstein1990HighAgreementLow,byrt1993BiasPrevalenceKappa}, which is relevant for both \textit{uses-evidence} and \textit{uses-knowledge} where most responses are labeled \textit{Yes} and \textit{No}, respectively.
So, we emphasize raw agreement and binarized agreement alongside $\kappa$.

Because our primary objective is comparative benchmarking, we focus on \emph{system-level} rank order rather than item-level absolute labels.
Consistent with this objective, system rankings are stable across aggregation methods (Pyramid vs.\ MACE) and remain similar under simulated smaller annotation budgets (Figure~\ref{fig:pyramid-mace-corr}).
That said, our study has limitations.
We did not conduct post-hoc consensus adjudication; future work could explore adjudication by practicing clinicians, although some residual disagreement is likely to remain because of ambiguity and differences in clinical judgment.
In addition, our annotators were trained medical students rather than independent clinicians.
While they had foundational clinical training, they did not have independent clinical experience.
They completed a calibration phase, and only annotators who met study criteria were included; however, their judgments may still differ from those of practicing clinicians.
This should be considered when interpreting the human reference labels and resulting system rankings.

Though our focus was on investigating the use of automated text and semantics metrics, there is an emerging strand of metrics on the rise, i.e., LLM-as-a-judge \cite{liu2023GEvalNLGEvaluation}.
However, there are mixed results in the healthcare domain, with some studies finding better correlations with human judgments \cite{croxford2025AutomatingEvaluationAI} while some found evidence of poor performance, underscoring the need for fine-tuning and adapting the LLMs before their use in evaluation as the judges \cite{laskar2025ImprovingAutomaticEvaluation}.
It is an interesting direction for further investigations into automated evaluation.

Decades of work in natural language processing (NLP) and medical AI show that progress follows measurement, i.e., we improve what we can evaluate.
By making evaluation rapid and reproducible, shared resources such as SQuAD \cite{rajpurkar2016SQuAD100000}, GLUE \cite{wang2018GLUEMultiTaskBenchmark}, and MedQA \cite{jin2021WhatDiseaseDoes} have shortened the build-measure-learn cycle, enabled system comparisons and targeted error analysis, and thereby driven successive performance gains; in contrast, reliance on labor‑intensive expert review slows iteration and limits scale.
In this context, we studied the automated evaluation of AI responses to hospitalization‑related patient questions along three clinically salient dimensions.
Grounded in large‑scale human annotations, our findings make rigorous assessment low‑cost, consistent, and near‑real‑time.
We believe our findings enable rapid benchmarking and error analysis, and provide a foundation for selecting and improving systems that support patient-clinician communication.

\section{Methods}

\subsection{Task}

We investigate the task of grounded QA over electronic health records (EHRs).
Specifically, for a given patient-posed question, its clinician-interpreted formulation, and a clinical note excerpt as evidence, the task is to return a 75-word natural language answer with specific citations to the key EHR note sentences.

\subsection{Data}

The patient cases in our evaluation are sampled from the ArchEHR-QA dataset \cite{soni2026DatasetAddressingPatients}, which was curated using real patient questions posted to public health forums about recent hospitalizations.
Each case consists of a question, a clinician's interpretation of the question, a clinical note excerpt with sentence-level relevance labels, and a clinician-authored free-text answer.
Questions were manually associated with representative clinical evidence from 
publicly accessible EHR databases (MIMIC-III and IV).
Subject matter experts (including clinicians) annotated the specific sentences from a clinical note that are essential, supplementary, or not relevant in answering the corresponding question.
For each question, clinicians also authored a 75-word natural language answer, which was grounded in the clinical note excerpt via citations to its specific sentences.
With the rich sentence-level relevance annotations and associated free-text answers, the dataset serves as a strong candidate for the experiments.

\subsection{System-generated Answers}

We collected system-generated answers for each question in our sample through a community evaluation at the Association for Computational Linguistics (ACL) 2025 conference \cite{soni2025OverviewArchEHRQA2025}.
For this study, we evaluate 28 end-to-end answer-generation systems.
Of these, 24 systems are shared task submissions summarized in the shared task overview paper \cite{soni2025OverviewArchEHRQA2025}, and the remaining four are baseline pipelines described in the ArchEHR-QA dataset paper \cite{soni2026DatasetAddressingPatients} (Together/Answer-First prompting strategies paired with Llama~4~17B~16E and Mixtral~8x22B models).
All systems generate answers in fewer than 75 words and provide sentence-level citations to supporting clinical note evidence.
Across systems, methodologies largely follow a multi-component pipeline that incorporates one or more of: evidence selection (selecting relevant evidence sentences from the note), answer generation (generating answer text), citation assignment (assigning or reassigning citations), and answer reformulation (rewriting the answer text).
To contextualize the diversity of system quality, Supplementary Table~S1 reports system-by-system results using the automated evaluation metrics.
Across all 28 systems, the overall factuality (strict citation F1 Score), overall relevance (average of all relevance scores using the clinician-authored answers as reference), and the overall score (mean of the overall factuality and relevance) span 20.2\%--62.3\%, 15.4\%--35.9\%, and 23.8\%--48.5\%, respectively.
Detailed results and system descriptions are available in the shared task overview and dataset papers \cite{soni2025OverviewArchEHRQA2025,soni2026DatasetAddressingPatients}.

\subsection{Automated Evaluation Metrics}

We used a suite of automated evaluation metrics to assess the system-generated answers.
These metrics broadly fall into two categories: \textit{Factuality} and \textit{Relevance}.
\textit{Factuality} measured citation precision/recall/F1 scores by comparing the evidence sentences cited in each system-generated answer to the gold sentence-level relevance labels (macro-averaged across answers).
We report two variants: a \textit{strict} variant, which treats only sentences labeled \textit{`essential'} as required facts, and a \textit{lenient} variant, which treats sentences labeled \textit{`essential`} or \textit{`supplementary`} as relevant.
\textit{Relevance} measured lexical or semantic similarity between the system-generated answer text and a reference text.
Text-based metrics included BLEU \cite{papineni2002BLEUMethodAutomatic}, ROUGE \cite{lin2004ROUGEPackageAutomatic}, and SARI \cite{xu2016OptimizingStatisticalMachine}, and semantics-based metrics included BERTScore \cite{zhang2019BERTScoreEvaluatingText}, AlignScore \cite{zha2023AlignScoreEvaluatingFactual}, and MEDCON \cite{yim2023AcibenchNovelAmbient}.
We computed \textit{Relevance} against two reference types: clinician-authored reference answers and a proxy reference formed by concatenating the note sentences labeled \textit{essential} with the input question (a scenario where relevance labels are available but clinician-authored answers are not).
We refer to \textit{Relevance} metrics computed against clinician-authored reference answers or the note-based proxy reference as ``(human)'' or ``(note)'', respectively.

\subsection{Human Judgments}

Rubric design focused on three clinically relevant facets of a grounded answer: (i) whether the response answers the patient question (\textit{answers-question}), (ii) whether answer assertions appropriately use and are consistent with the provided clinical note evidence (\textit{uses-evidence}), and (iii) whether any additional general medical knowledge introduced beyond the note is appropriate (\textit{uses-knowledge}).
This multi-axis structure follows established evaluation practice for evidence-grounded text generation, such as in the TREC BioGen and RAG tracks \cite{gupta2024OverviewTREC2024,thakur2025AssessingSupportTREC}.
Table~\ref{tab:evaluation-dimensions} summarizes the definitions and label sets used for each dimension.
Full annotation guidelines with additional examples are provided as Supplementary Materials.

Annotator training included a calibration phase with physician review and feedback on early assignments.
A pool of $26$ medical students served as annotators.
Each system response was independently rated by three different annotators to reduce bias.
Annotators completed a calibration phase, during which each annotator's ratings for the first five system responses were reviewed by a physician, and feedback was provided to correct any misunderstandings of the rubric.
If the annotators made obvious mistakes during calibration, these were corrected, but any legitimate differences in opinion were retained for this open-ended task.
After calibration, annotators worked independently and did not discuss or adjudicate individual items.
The agreement statistics in Table~\ref{tab:agreement} reflect pre-adjudication disagreement.

Label scoring and aggregation supported system-level ranking by mapping each categorical judgment to a numerical score.
For the \textit{answers-question} dimension, we map \textit{Yes} to $2$, \textit{Partially} to $1$, and \textit{No} to $0$.
For \textit{uses-evidence} and \textit{uses-knowledge}, we map \textit{Yes} to $1$, \textit{No} to $0$, and \textit{Refutes}/\textit{Conflicting} to $-1$.
We then aggregate annotations for system ranking using two approaches: \textit{Pyramid} (summing the mapped scores across annotators per case) \cite{nenkova2004EvaluatingContentSelection} and \textit{MACE} (Multi-Annotator Competence Estimation), an unsupervised model that infers latent labels while estimating annotator reliability and provides confidence-scored predictions that often outperform majority voting \cite{hovy2013LearningWhomTrust}.

\subsection{Evaluation}

We report inter-annotator agreements to highlight the task complexity and labeling quality.
We also correlate the scores achieved using the different score accumulation techniques.
We rank the systems based on all the automated metrics and human-labeled scores individually.
We correlate the system rankings across different metrics and report the Kendall's Tau correlation coefficients.

\section*{Data Availability}

The ArchEHR-QA dataset, which provided the patient cases for this evaluation study, is publicly available on PhysioNet at \url{https://doi.org/10.13026/zzax-sy62}.

\section*{Code Availability}

The accompanying code used to evaluate the system responses is available via GitHub at \url{https://github.com/soni-sarvesh/archehr-qa/tree/main/evaluation}.

\section*{Acknowledgments}

This research was supported by the Intramural Research Program of the National Institutes of Health (NIH) and utilized the computational resources of the NIH HPC Biowulf cluster (http://hpc.nih.gov). The contributions of the NIH authors are considered Works of the United States Government. The findings and conclusions presented in this paper are those of the authors and do not necessarily reflect the views of the NIH or the U.S. Department of Health and Human Services.

\bibliographystyle{unsrt}  
\bibliography{references}  

@article{ayers2023ComparingPhysicianArtificial,
  title = {Comparing {{Physician}} and {{Artificial Intelligence Chatbot Responses}} to {{Patient Questions Posted}} to a {{Public Social Media Forum}}},
  author = {Ayers, John W. and Poliak, Adam and Dredze, Mark and Leas, Eric C. and Zhu, Zechariah and Kelley, Jessica B. and Faix, Dennis J. and Goodman, Aaron M. and Longhurst, Christopher A. and Hogarth, Michael and Smith, Davey M.},
  year = {2023},
  month = jun,
  journal = {JAMA Internal Medicine},
  volume = {183},
  number = {6},
  pages = {589--596},
  issn = {2168-6106},
  doi = {10.1001/jamainternmed.2023.1838},
  urldate = {2024-09-26}
}

@inproceedings{balepur2024ArtifactsAbductionHow,
  title = {Artifacts or {{Abduction}}: {{How Do LLMs Answer Multiple-Choice Questions Without}} the {{Question}}?},
  shorttitle = {Artifacts or {{Abduction}}},
  booktitle = {Proceedings of the 62nd {{Annual Meeting}} of the {{Association}} for {{Computational Linguistics}} ({{Volume}} 1: {{Long Papers}})},
  author = {Balepur, Nishant and Ravichander, Abhilasha and Rudinger, Rachel},
  editor = {Ku, Lun-Wei and Martins, Andre and Srikumar, Vivek},
  year = {2024},
  month = aug,
  pages = {10308--10330},
  publisher = {Association for Computational Linguistics},
  address = {Bangkok, Thailand},
  doi = {10.18653/v1/2024.acl-long.555},
  urldate = {2025-09-12}
}

@inproceedings{bartels2025CanLargeLanguage,
  title = {Can {{Large Language Models Accurately Generate Answer Keys}} for {{Health-related Questions}}?},
  booktitle = {Proceedings of the 63rd {{Annual Meeting}} of the {{Association}} for {{Computational Linguistics}} ({{Volume}} 2: {{Short Papers}})},
  author = {Bartels, Davis and Gupta, Deepak and {Demner-Fushman}, Dina},
  editor = {Che, Wanxiang and Nabende, Joyce and Shutova, Ekaterina and Pilehvar, Mohammad Taher},
  year = {2025},
  month = jul,
  pages = {354--368},
  publisher = {Association for Computational Linguistics},
  address = {Vienna, Austria},
  doi = {10.18653/v1/2025.acl-short.28},
  urldate = {2025-09-15},
  isbn = {979-8-89176-252-7}
}

@article{bedi2025TestingEvaluationHealth,
  title = {Testing and {{Evaluation}} of {{Health Care Applications}} of {{Large Language Models}}: {{A Systematic Review}}},
  shorttitle = {Testing and {{Evaluation}} of {{Health Care Applications}} of {{Large Language Models}}},
  author = {Bedi, Suhana and Liu, Yutong and {Orr-Ewing}, Lucy and Dash, Dev and Koyejo, Sanmi and Callahan, Alison and Fries, Jason A. and Wornow, Michael and Swaminathan, Akshay and Lehmann, Lisa Soleymani and Hong, Hyo Jung and Kashyap, Mehr and Chaurasia, Akash R. and Shah, Nirav R. and Singh, Karandeep and Tazbaz, Troy and Milstein, Arnold and Pfeffer, Michael A. and Shah, Nigam H.},
  year = {2025},
  month = jan,
  journal = {JAMA},
  volume = {333},
  number = {4},
  pages = {319--328},
  issn = {1538-3598},
  doi = {10.1001/jama.2024.21700},
  langid = {english},
  pmcid = {PMC11480901},
  pmid = {39405325}
}

@article{bernstein2023ComparisonOphthalmologistLarge,
  title = {Comparison of {{Ophthalmologist}} and {{Large Language Model Chatbot Responses}} to {{Online Patient Eye Care Questions}}},
  author = {Bernstein, Isaac A. and Zhang, Youchen (Victor) and Govil, Devendra and Majid, Iyad and Chang, Robert T. and Sun, Yang and Shue, Ann and Chou, Jonathan C. and Schehlein, Emily and Christopher, Karen L. and Groth, Sylvia L. and Ludwig, Cassie and Wang, Sophia Y.},
  year = {2023},
  month = aug,
  journal = {JAMA Network Open},
  volume = {6},
  number = {8},
  pages = {e2330320},
  issn = {2574-3805},
  doi = {10.1001/jamanetworkopen.2023.30320},
  urldate = {2024-09-26}
}

@article{busch2025CurrentApplicationsChallenges,
  title = {Current Applications and Challenges in Large Language Models for Patient Care: A Systematic Review},
  shorttitle = {Current Applications and Challenges in Large Language Models for Patient Care},
  author = {Busch, Felix and Hoffmann, Lena and Rueger, Christopher and {van Dijk}, Elon HC and Kader, Rawen and {Ortiz-Prado}, Esteban and Makowski, Marcus R. and Saba, Luca and Hadamitzky, Martin and Kather, Jakob Nikolas and Truhn, Daniel and Cuocolo, Renato and Adams, Lisa C. and Bressem, Keno K.},
  year = {2025},
  month = jan,
  journal = {Communications Medicine},
  volume = {5},
  number = {1},
  pages = {26},
  publisher = {Nature Publishing Group},
  issn = {2730-664X},
  doi = {10.1038/s43856-024-00717-2},
  urldate = {2025-09-11},
  copyright = {2025 The Author(s)},
  langid = {english}
}

@inproceedings{chen2024MLLMasaJudgeAssessingMultimodal,
  title = {{{MLLM-as-a-Judge}}: {{Assessing Multimodal LLM-as-a-Judge}} with {{Vision-Language Benchmark}}},
  shorttitle = {{{MLLM-as-a-Judge}}},
  booktitle = {Proceedings of the 41st {{International Conference}} on {{Machine Learning}}},
  author = {Chen, Dongping and Chen, Ruoxi and Zhang, Shilin and Wang, Yaochen and Liu, Yinuo and Zhou, Huichi and Zhang, Qihui and Wan, Yao and Zhou, Pan and Sun, Lichao},
  year = {2024},
  month = jul,
  pages = {6562--6595},
  publisher = {PMLR},
  issn = {2640-3498},
  urldate = {2025-09-16},
  langid = {english}
}

@misc{croxford2025AutomatingEvaluationAI,
  title = {Automating {{Evaluation}} of {{AI Text Generation}} in {{Healthcare}} with a {{Large Language Model}} ({{LLM}})-as-a-{{Judge}}},
  author = {Croxford, Emma and Gao, Yanjun and First, Elliot and Pellegrino, Nicholas and Schnier, Miranda and Caskey, John and Oguss, Madeline and Wills, Graham and Chen, Guanhua and Dligach, Dmitriy and Churpek, Matthew M. and Mayampurath, Anoop and Liao, Frank and Goswami, Cherodeep and Wong, Karen K. and Patterson, Brian W. and Afshar, Majid},
  year = {2025},
  month = may,
  pages = {2025.04.22.25326219},
  publisher = {medRxiv},
  doi = {10.1101/2025.04.22.25326219},
  urldate = {2025-09-22},
  archiveprefix = {medRxiv},
  copyright = {{\copyright} 2025, Posted by Cold Spring Harbor Laboratory. This pre-print is available under a Creative Commons License (Attribution 4.0 International), CC BY 4.0, as described at http://creativecommons.org/licenses/by/4.0/},
  langid = {english}
}

@article{farea2025EvaluationQuestionAnswering,
  title = {Evaluation of {{Question Answering Systems}}: {{Complexity}} of {{Judging}} a {{Natural Language}}},
  shorttitle = {Evaluation of {{Question Answering Systems}}},
  author = {Farea, Amer and Yang, Zhen and Duong, Kien and Perera, Nadeesha and {Emmert-Streib}, Frank},
  year = {2025},
  month = aug,
  journal = {ACM Comput. Surv.},
  volume = {58},
  number = {1},
  pages = {1:1--1:43},
  issn = {0360-0300},
  doi = {10.1145/3744663},
  urldate = {2025-09-15}
}

@inproceedings{griot2025PatternRecognitionMedical,
  title = {Pattern {{Recognition}} or {{Medical Knowledge}}? {{The Problem}} with {{Multiple-Choice Questions}} in {{Medicine}}},
  shorttitle = {Pattern {{Recognition}} or {{Medical Knowledge}}?},
  booktitle = {Proceedings of the 63rd {{Annual Meeting}} of the {{Association}} for {{Computational Linguistics}} ({{Volume}} 1: {{Long Papers}})},
  author = {Griot, Maxime and Vanderdonckt, Jean and Yuksel, Demet and Hemptinne, Coralie},
  editor = {Che, Wanxiang and Nabende, Joyce and Shutova, Ekaterina and Pilehvar, Mohammad Taher},
  year = {2025},
  month = jul,
  pages = {5321--5341},
  publisher = {Association for Computational Linguistics},
  address = {Vienna, Austria},
  doi = {10.18653/v1/2025.acl-long.266},
  urldate = {2025-09-12},
  isbn = {979-8-89176-251-0}
}

@inproceedings{hanna2021FineGrainedAnalysisBERTScore,
  title = {A {{Fine-Grained Analysis}} of {{BERTScore}}},
  booktitle = {Proceedings of the {{Sixth Conference}} on {{Machine Translation}}},
  author = {Hanna, Michael and Bojar, Ond{\v r}ej},
  editor = {Barrault, Loic and Bojar, Ondrej and Bougares, Fethi and Chatterjee, Rajen and {Costa-jussa}, Marta R. and Federmann, Christian and Fishel, Mark and Fraser, Alexander and Freitag, Markus and Graham, Yvette and Grundkiewicz, Roman and Guzman, Paco and Haddow, Barry and Huck, Matthias and Yepes, Antonio Jimeno and Koehn, Philipp and Kocmi, Tom and Martins, Andre and Morishita, Makoto and Monz, Christof},
  year = {2021},
  month = nov,
  pages = {507--517},
  publisher = {Association for Computational Linguistics},
  address = {Online},
  urldate = {2025-09-21}
}

@inproceedings{hovy2013LearningWhomTrust,
  title = {Learning {{Whom}} to {{Trust}} with {{MACE}}},
  booktitle = {Proceedings of the 2013 {{Conference}} of the {{North American Chapter}} of the {{Association}} for {{Computational Linguistics}}: {{Human Language Technologies}}},
  author = {Hovy, Dirk and {Berg-Kirkpatrick}, Taylor and Vaswani, Ashish and Hovy, Eduard},
  editor = {Vanderwende, Lucy and Daum{\'e} III, Hal and Kirchhoff, Katrin},
  year = {2013},
  month = jun,
  pages = {1120--1130},
  publisher = {Association for Computational Linguistics},
  address = {Atlanta, Georgia},
  urldate = {2025-09-18}
}

@article{jin2021WhatDiseaseDoes,
  title = {What {{Disease Does This Patient Have}}? {{A Large-Scale Open Domain Question Answering Dataset}} from {{Medical Exams}}},
  shorttitle = {What {{Disease Does This Patient Have}}?},
  author = {Jin, Di and Pan, Eileen and Oufattole, Nassim and Weng, Wei-Hung and Fang, Hanyi and Szolovits, Peter},
  year = {2021},
  month = jan,
  journal = {Applied Sciences},
  volume = {11},
  number = {14},
  pages = {6421},
  publisher = {Multidisciplinary Digital Publishing Institute},
  issn = {2076-3417},
  doi = {10.3390/app11146421},
  urldate = {2025-09-24},
  copyright = {http://creativecommons.org/licenses/by/3.0/},
  langid = {english}
}

@article{kung2023PerformanceChatGPTUSMLE,
  title = {Performance of {{ChatGPT}} on {{USMLE}}: {{Potential}} for {{AI-assisted}} Medical Education Using Large Language Models},
  shorttitle = {Performance of {{ChatGPT}} on {{USMLE}}},
  author = {Kung, Tiffany H. and Cheatham, Morgan and Medenilla, Arielle and Sillos, Czarina and Leon, Lorie De and Elepa{\~n}o, Camille and Madriaga, Maria and Aggabao, Rimel and {Diaz-Candido}, Giezel and Maningo, James and Tseng, Victor},
  year = {2023},
  month = feb,
  journal = {PLOS Digital Health},
  volume = {2},
  number = {2},
  pages = {e0000198},
  publisher = {Public Library of Science},
  issn = {2767-3170},
  doi = {10.1371/journal.pdig.0000198},
  urldate = {2025-09-12},
  langid = {english}
}

@inproceedings{laskar2025ImprovingAutomaticEvaluation,
  title = {Improving {{Automatic Evaluation}} of {{Large Language Models}} ({{LLMs}}) in {{Biomedical Relation Extraction}} via {{LLMs-as-the-Judge}}},
  booktitle = {Proceedings of the 63rd {{Annual Meeting}} of the {{Association}} for {{Computational Linguistics}} ({{Volume}} 1: {{Long Papers}})},
  author = {Laskar, Md Tahmid Rahman and Jahan, Israt and Dolatabadi, Elham and Peng, Chun and Hoque, Enamul and Huang, Jimmy},
  editor = {Che, Wanxiang and Nabende, Joyce and Shutova, Ekaterina and Pilehvar, Mohammad Taher},
  year = {2025},
  month = jul,
  pages = {25483--25497},
  publisher = {Association for Computational Linguistics},
  address = {Vienna, Austria},
  doi = {10.18653/v1/2025.acl-long.1238},
  urldate = {2025-09-22},
  isbn = {979-8-89176-251-0}
}

@inproceedings{li2024CanMultiplechoiceQuestions,
  title = {Can {{Multiple-choice Questions Really Be Useful}} in {{Detecting}} the {{Abilities}} of {{LLMs}}?},
  booktitle = {Proceedings of the 2024 {{Joint International Conference}} on {{Computational Linguistics}}, {{Language Resources}} and {{Evaluation}} ({{LREC-COLING}} 2024)},
  author = {Li, Wangyue and Li, Liangzhi and Xiang, Tong and Liu, Xiao and Deng, Wei and Garcia, Noa},
  editor = {Calzolari, Nicoletta and Kan, Min-Yen and Hoste, Veronique and Lenci, Alessandro and Sakti, Sakriani and Xue, Nianwen},
  year = {2024},
  month = may,
  pages = {2819--2834},
  publisher = {{ELRA and ICCL}},
  address = {Torino, Italia},
  urldate = {2025-09-12}
}

@inproceedings{lin2004ROUGEPackageAutomatic,
  title = {{{ROUGE}}: {{A Package}} for {{Automatic Evaluation}} of {{Summaries}}},
  shorttitle = {{{ROUGE}}},
  booktitle = {Text {{Summarization Branches Out}}},
  author = {Lin, Chin-Yew},
  year = {2004},
  month = jul,
  pages = {74--81},
  publisher = {Association for Computational Linguistics},
  address = {Barcelona, Spain},
  urldate = {2024-03-15}
}

@inproceedings{liu2023GEvalNLGEvaluation,
  title = {G-{{Eval}}: {{NLG Evaluation}} Using {{Gpt-4}} with {{Better Human Alignment}}},
  shorttitle = {G-{{Eval}}},
  booktitle = {Proceedings of the 2023 {{Conference}} on {{Empirical Methods}} in {{Natural Language Processing}}},
  author = {Liu, Yang and Iter, Dan and Xu, Yichong and Wang, Shuohang and Xu, Ruochen and Zhu, Chenguang},
  editor = {Bouamor, Houda and Pino, Juan and Bali, Kalika},
  year = {2023},
  month = dec,
  pages = {2511--2522},
  publisher = {Association for Computational Linguistics},
  address = {Singapore},
  doi = {10.18653/v1/2023.emnlp-main.153},
  urldate = {2025-09-22}
}

@article{liu2025ApplicationLargeLanguage,
  title = {Application of Large Language Models in Medicine},
  author = {Liu, Fenglin and Zhou, Hongjian and Gu, Boyang and Zou, Xinyu and Huang, Jinfa and Wu, Jinge and Li, Yiru and Chen, Sam S. and Hua, Yining and Zhou, Peilin and Liu, Junling and Mao, Chengfeng and You, Chenyu and Wu, Xian and Zheng, Yefeng and Clifton, Lei and Li, Zheng and Luo, Jiebo and Clifton, David A.},
  year = {2025},
  month = jun,
  journal = {Nature Reviews Bioengineering},
  volume = {3},
  number = {6},
  pages = {445--464},
  publisher = {Nature Publishing Group},
  issn = {2731-6092},
  doi = {10.1038/s44222-025-00279-5},
  urldate = {2025-09-13},
  copyright = {2025  Springer Nature Limited},
  langid = {english}
}

@inproceedings{marton2006NuggeteerAutomaticNuggetBased,
  title = {Nuggeteer: {{Automatic Nugget-Based Evaluation}} Using {{Descriptions}} and {{Judgements}}},
  shorttitle = {Nuggeteer},
  booktitle = {Proceedings of the {{Human Language Technology Conference}} of the {{NAACL}}, {{Main Conference}}},
  author = {Marton, Gregory and Radul, Alexey},
  editor = {Moore, Robert C. and Bilmes, Jeff and {Chu-Carroll}, Jennifer and Sanderson, Mark},
  year = {2006},
  month = jun,
  pages = {375--382},
  publisher = {Association for Computational Linguistics},
  address = {New York City, USA},
  urldate = {2025-09-15}
}

@inproceedings{moramarco2022HumanEvaluationCorrelation,
  title = {Human {{Evaluation}} and {{Correlation}} with {{Automatic Metrics}} in {{Consultation Note Generation}}},
  booktitle = {Proceedings of the 60th {{Annual Meeting}} of the {{Association}} for {{Computational Linguistics}} ({{Volume}} 1: {{Long Papers}})},
  author = {Moramarco, Francesco and Papadopoulos Korfiatis, Alex and Perera, Mark and Juric, Damir and Flann, Jack and Reiter, Ehud and Belz, Anya and Savkov, Aleksandar},
  year = {2022},
  month = may,
  pages = {5739--5754},
  publisher = {Association for Computational Linguistics},
  address = {Dublin, Ireland},
  doi = {10.18653/v1/2022.acl-long.394},
  urldate = {2023-08-18}
}

@article{naveed2025ComprehensiveOverviewLarge,
  title = {A {{Comprehensive Overview}} of {{Large Language Models}}},
  author = {Naveed, Humza and Khan, Asad Ullah and Qiu, Shi and Saqib, Muhammad and Anwar, Saeed and Usman, Muhammad and Akhtar, Naveed and Barnes, Nick and Mian, Ajmal},
  year = {2025},
  month = aug,
  journal = {ACM Trans. Intell. Syst. Technol.},
  volume = {16},
  number = {5},
  pages = {106:1--106:72},
  issn = {2157-6904},
  doi = {10.1145/3744746},
  urldate = {2025-09-13}
}

@inproceedings{nenkova2004EvaluatingContentSelection,
  title = {Evaluating {{Content Selection}} in {{Summarization}}: {{The Pyramid Method}}},
  shorttitle = {Evaluating {{Content Selection}} in {{Summarization}}},
  booktitle = {Proceedings of the {{Human Language Technology Conference}} of the {{North American Chapter}} of the {{Association}} for {{Computational Linguistics}}: {{HLT-NAACL}} 2004},
  author = {Nenkova, Ani and Passonneau, Rebecca},
  year = {2004},
  month = may,
  pages = {145--152},
  publisher = {Association for Computational Linguistics},
  address = {Boston, Massachusetts, USA},
  urldate = {2025-09-15}
}

@inproceedings{nguyen2024ReferencebasedMetricsDisprove,
  title = {Reference-Based {{Metrics Disprove Themselves}} in {{Question Generation}}},
  booktitle = {Findings of the {{Association}} for {{Computational Linguistics}}: {{EMNLP}} 2024},
  author = {Nguyen, Bang and Yu, Mengxia and Huang, Yun and Jiang, Meng},
  editor = {{Al-Onaizan}, Yaser and Bansal, Mohit and Chen, Yun-Nung},
  year = {2024},
  month = nov,
  pages = {13651--13666},
  publisher = {Association for Computational Linguistics},
  address = {Miami, Florida, USA},
  doi = {10.18653/v1/2024.findings-emnlp.798},
  urldate = {2025-09-21}
}

@inproceedings{novikova2017WhyWeNeed,
  title = {Why {{We Need New Evaluation Metrics}} for {{NLG}}},
  booktitle = {Proceedings of the 2017 {{Conference}} on {{Empirical Methods}} in {{Natural Language Processing}}},
  author = {Novikova, Jekaterina and Du{\v s}ek, Ond{\v r}ej and Cercas Curry, Amanda and Rieser, Verena},
  editor = {Palmer, Martha and Hwa, Rebecca and Riedel, Sebastian},
  year = {2017},
  month = sep,
  pages = {2241--2252},
  publisher = {Association for Computational Linguistics},
  address = {Copenhagen, Denmark},
  doi = {10.18653/v1/D17-1238},
  urldate = {2025-09-21}
}

@article{osnat2025PatientPerspectivesArtificial,
  title = {Patient Perspectives on Artificial Intelligence in Healthcare: {{A}} Global Scoping Review of Benefits, Ethical Concerns, and Implementation Strategies},
  shorttitle = {Patient Perspectives on Artificial Intelligence in Healthcare},
  author = {Osnat, Bashkin},
  year = {2025},
  month = nov,
  journal = {International Journal of Medical Informatics},
  volume = {203},
  pages = {106007},
  issn = {1386-5056},
  doi = {10.1016/j.ijmedinf.2025.106007},
  urldate = {2025-09-12}
}

@inproceedings{papineni2002BLEUMethodAutomatic,
  title = {{{BLEU}}: A Method for Automatic Evaluation of Machine Translation},
  booktitle = {Proceedings of the 40th Annual Meeting on Association for Computational Linguistics},
  author = {Papineni, Kishore and Roukos, Salim and Ward, Todd and Zhu, Wei-Jing},
  year = {2002},
  pages = {311--318},
  publisher = {Association for Computational Linguistics},
  doi = {10.3115/1073083.1073135}
}

@inproceedings{rajpurkar2016SQuAD100000,
  title = {{{SQuAD}}: 100,000+ {{Questions}} for {{Machine Comprehension}} of {{Text}}},
  shorttitle = {{{SQuAD}}},
  booktitle = {Proceedings of the 2016 {{Conference}} on {{Empirical Methods}} in {{Natural Language Processing}}},
  author = {Rajpurkar, Pranav and Zhang, Jian and Lopyrev, Konstantin and Liang, Percy},
  year = {2016},
  month = nov,
  pages = {2383--2392},
  publisher = {Association for Computational Linguistics},
  address = {Austin, Texas},
  doi = {10.18653/v1/D16-1264},
  urldate = {2019-11-22}
}

@article{sai2022SurveyEvaluationMetrics,
  title = {A {{Survey}} of {{Evaluation Metrics Used}} for {{NLG Systems}}},
  author = {Sai, Ananya B. and Mohankumar, Akash Kumar and Khapra, Mitesh M.},
  year = {2022},
  month = jan,
  journal = {ACM Comput. Surv.},
  volume = {55},
  number = {2},
  pages = {26:1--26:39},
  issn = {0360-0300},
  doi = {10.1145/3485766},
  urldate = {2025-09-16}
}

@article{shahsavar2023UserIntentionsUse,
  title = {User {{Intentions}} to {{Use ChatGPT}} for {{Self-Diagnosis}} and {{Health-Related Purposes}}: {{Cross-sectional Survey Study}}},
  shorttitle = {User {{Intentions}} to {{Use ChatGPT}} for {{Self-Diagnosis}} and {{Health-Related Purposes}}},
  author = {Shahsavar, Yeganeh and Choudhury, Avishek},
  year = {2023},
  month = may,
  journal = {JMIR Human Factors},
  volume = {10},
  number = {1},
  pages = {e47564},
  publisher = {JMIR Publications Inc., Toronto, Canada},
  doi = {10.2196/47564},
  urldate = {2025-09-12},
  copyright = {This is an open-access article distributed under the terms of the Creative Commons Attribution License (https://creativecommons.org/licenses/by/4.0/), which permits unrestricted use, distribution, and reproduction in any medium, provided the original work, first published JMIR Human Factors, is properly cited. The complete bibliographic information, a link to the original publication on https://humanfactors.jmir.org/, as well as this copyright and license information must be included.},
  langid = {english}
}

@article{soni2026DatasetAddressingPatients,
  title = {A {{Dataset}} for {{Addressing Patient}}'s {{Information Needs}} Related to {{Clinical Course}} of {{Hospitalization}}},
  author = {Soni, Sarvesh and {Demner-Fushman}, Dina},
  year = 2026,
  month = feb,
  journal = {Scientific Data},
  volume = {13},
  number = {1},
  eprint = {2506.04156},
  primaryclass = {cs},
  pages = {523},
  publisher = {Nature Publishing Group},
  issn = {2052-4463},
  doi = {10.1038/s41597-026-06639-z},
  urldate = {2026-05-07},
  archiveprefix = {arXiv},
  copyright = {2026 This is a U.S. Government work and not under copyright protection in the US; foreign copyright protection may apply},
  langid = {english},
  pmcid = {PMC13046737},
  pmid = {41741485}
}

@inproceedings{soni2025OverviewArchEHRQA2025,
  title = {Overview of the {{ArchEHR-QA}} 2025 {{Shared Task}} on {{Grounded Question Answering}} from {{Electronic Health Records}}},
  booktitle = {Proceedings of the 24th {{Workshop}} on {{Biomedical Language Processing}}},
  author = {Soni, Sarvesh and Gayen, Soumya and {Demner-Fushman}, Dina},
  editor = {{Demner-Fushman}, Dina and Ananiadou, Sophia and Miwa, Makoto and Tsujii, Junichi},
  year = {2025},
  month = aug,
  pages = {396--405},
  publisher = {Association for Computational Linguistics},
  address = {Viena, Austria},
  urldate = {2025-07-25},
  isbn = {979-8-89176-275-6}
}

@inproceedings{wang2018GLUEMultiTaskBenchmark,
  title = {{{GLUE}}: {{A Multi-Task Benchmark}} and {{Analysis Platform}} for {{Natural Language Understanding}}},
  shorttitle = {{{GLUE}}},
  booktitle = {Proceedings of the 2018 {{EMNLP Workshop BlackboxNLP}}: {{Analyzing}} and {{Interpreting Neural Networks}} for {{NLP}}},
  author = {Wang, Alex and Singh, Amanpreet and Michael, Julian and Hill, Felix and Levy, Omer and Bowman, Samuel},
  editor = {Linzen, Tal and Chrupa{\l}a, Grzegorz and Alishahi, Afra},
  year = {2018},
  month = nov,
  pages = {353--355},
  publisher = {Association for Computational Linguistics},
  address = {Brussels, Belgium},
  doi = {10.18653/v1/W18-5446},
  urldate = {2025-09-24}
}

@article{xu2016OptimizingStatisticalMachine,
  title = {Optimizing {{Statistical Machine Translation}} for {{Text Simplification}}},
  author = {Xu, Wei and Napoles, Courtney and Pavlick, Ellie and Chen, Quanze and {Callison-Burch}, Chris},
  editor = {Lee, Lillian and Johnson, Mark and Toutanova, Kristina},
  year = {2016},
  journal = {Transactions of the Association for Computational Linguistics},
  volume = {4},
  pages = {401--415},
  publisher = {MIT Press},
  address = {Cambridge, MA},
  doi = {10.1162/tacl_a_00107},
  urldate = {2025-05-31}
}

@article{yim2023AcibenchNovelAmbient,
  title = {Aci-Bench: A {{Novel Ambient Clinical Intelligence Dataset}} for {{Benchmarking Automatic Visit Note Generation}}},
  shorttitle = {Aci-Bench},
  author = {Yim, Wen-wai and Fu, Yujuan and Ben Abacha, Asma and Snider, Neal and Lin, Thomas and Yetisgen, Meliha},
  year = {2023},
  month = sep,
  journal = {Scientific Data},
  volume = {10},
  number = {1},
  pages = {586},
  publisher = {Nature Publishing Group},
  issn = {2052-4463},
  doi = {10.1038/s41597-023-02487-3},
  urldate = {2023-10-15},
  copyright = {2023 Springer Nature Limited},
  langid = {english}
}

@inproceedings{zha2023AlignScoreEvaluatingFactual,
  title = {{{AlignScore}}: {{Evaluating Factual Consistency}} with {{A Unified Alignment Function}}},
  shorttitle = {{{AlignScore}}},
  booktitle = {Proceedings of the 61st {{Annual Meeting}} of the {{Association}} for {{Computational Linguistics}} ({{Volume}} 1: {{Long Papers}})},
  author = {Zha, Yuheng and Yang, Yichi and Li, Ruichen and Hu, Zhiting},
  editor = {Rogers, Anna and {Boyd-Graber}, Jordan and Okazaki, Naoaki},
  year = {2023},
  month = jul,
  pages = {11328--11348},
  publisher = {Association for Computational Linguistics},
  address = {Toronto, Canada},
  doi = {10.18653/v1/2023.acl-long.634},
  urldate = {2025-01-16}
}

@inproceedings{zhang2019BERTScoreEvaluatingText,
  title = {{{BERTScore}}: {{Evaluating Text Generation}} with {{BERT}}},
  shorttitle = {{{BERTScore}}},
  booktitle = {International {{Conference}} on {{Learning Representations}}},
  author = {Zhang, Tianyi and Kishore, Varsha and Wu, Felix and Weinberger, Kilian Q. and Artzi, Yoav},
  year = {2019},
  month = sep,
  urldate = {2024-03-10},
  langid = {english}
}

@misc{gupta2024OverviewTREC2024,
  title = {Overview of {{TREC}} 2024 {{Biomedical Generative Retrieval}} ({{BioGen}}) {{Track}}},
  author = {Gupta, Deepak and {Demner-Fushman}, Dina and Hersh, William and Bedrick, Steven and Roberts, Kirk},
  year = 2024,
  month = dec,
  number = {arXiv:2411.18069},
  eprint = {2411.18069},
  primaryclass = {cs},
  publisher = {arXiv},
  doi = {10.48550/arXiv.2411.18069},
  urldate = {2026-02-06},
  archiveprefix = {arXiv}
}

@inproceedings{thakur2025AssessingSupportTREC,
  title = {Assessing {{Support}} for the {{TREC}} 2024 {{RAG Track}}: {{A Large-Scale Comparative Study}} of {{LLM}} and {{Human Evaluations}}},
  shorttitle = {Assessing {{Support}} for the {{TREC}} 2024 {{RAG Track}}},
  booktitle = {Proceedings of the 48th {{International ACM SIGIR Conference}} on {{Research}} and {{Development}} in {{Information Retrieval}}},
  author = {Thakur, Nandan and Pradeep, Ronak and Upadhyay, Shivani and Campos, Daniel and Craswell, Nick and Soboroff, Ian and Dang, Hoa Trang and Lin, Jimmy},
  year = 2025,
  month = jul,
  series = {{{SIGIR}} '25},
  pages = {2759--2763},
  publisher = {Association for Computing Machinery},
  address = {New York, NY, USA},
  doi = {10.1145/3726302.3730165},
  urldate = {2026-02-06},
  isbn = {979-8-4007-1592-1}
}

@article{byrt1993BiasPrevalenceKappa,
  title = {Bias, Prevalence and Kappa},
  author = {Byrt, Ted and Bishop, Janet and Carlin, John B.},
  year = 1993,
  month = may,
  journal = {Journal of Clinical Epidemiology},
  volume = {46},
  number = {5},
  pages = {423--429},
  issn = {0895-4356},
  doi = {10.1016/0895-4356(93)90018-V},
  urldate = {2026-02-06}
}

@article{feinstein1990HighAgreementLow,
  title = {High Agreement but Low {{Kappa}}: {{I}}. the Problems of Two Paradoxes},
  shorttitle = {High Agreement but Low {{Kappa}}},
  author = {Feinstein, Alvan R. and Cicchetti, Domenic V.},
  year = 1990,
  month = jan,
  journal = {Journal of Clinical Epidemiology},
  volume = {43},
  number = {6},
  pages = {543--549},
  issn = {0895-4356},
  doi = {10.1016/0895-4356(90)90158-L},
  urldate = {2026-02-06}
}

@article{fernandez-llimos2025ConsistencyMedicalSubject,
  title = {Consistency of {{Medical Subject Headings}} Assignment: {{A}} Test-Retest Reliability Analysis},
  shorttitle = {Consistency of {{Medical Subject Headings}} Assignment},
  author = {{Fernandez-Llimos}, Fernando},
  year = 2025,
  month = oct,
  journal = {Research in Social and Administrative Pharmacy},
  volume = {21},
  number = {10},
  pages = {784--789},
  issn = {1551-7411},
  doi = {10.1016/j.sapharm.2025.05.008},
  urldate = {2026-02-06}
}

@article{funk1983IndexingConsistencyMEDLINE,
  title = {Indexing Consistency in {{MEDLINE}}},
  author = {Funk, M. E. and Reid, C. A.},
  year = 1983,
  month = apr,
  journal = {Bulletin of the Medical Library Association},
  volume = {71},
  number = {2},
  pages = {176},
  urldate = {2026-02-06},
  langid = {english},
  pmid = {6344946}
}

@article{novack2006DisagreementInterpretationChest,
  title = {Disagreement in the Interpretation of Chest Radiographs among Specialists and Clinical Outcomes of Patients Hospitalized with Suspected Pneumonia},
  author = {Novack, Victor and Avnon, Lone S. and Smolyakov, Alexander and Barnea, Rachel and Jotkowitz, Alan and Schlaeffer, Francisc},
  year = 2006,
  month = jan,
  journal = {European Journal of Internal Medicine},
  volume = {17},
  number = {1},
  pages = {43--47},
  issn = {0953-6205},
  doi = {10.1016/j.ejim.2005.07.008},
  urldate = {2026-02-06}
}

@inproceedings{benabacha2023EmpiricalStudyClinical,
  title = {An {{Empirical Study}} of {{Clinical Note Generation}} from {{Doctor-Patient Encounters}}},
  booktitle = {Proceedings of the 17th {{Conference}} of the {{European Chapter}} of the {{Association}} for {{Computational Linguistics}}},
  author = {Ben Abacha, Asma and Yim, Wen-wai and Fan, Yadan and Lin, Thomas},
  year = 2023,
  month = may,
  pages = {2291--2302},
  publisher = {Association for Computational Linguistics},
  address = {Dubrovnik, Croatia},
  urldate = {2023-08-21}
}

@inproceedings{benabacha2023InvestigationEvaluationMethods,
  title = {An {{Investigation}} of {{Evaluation Methods}} in {{Automatic Medical Note Generation}}},
  booktitle = {Findings of the {{Association}} for {{Computational Linguistics}}: {{ACL}} 2023},
  author = {Ben Abacha, Asma and Yim, Wen-wai and Michalopoulos, George and Lin, Thomas},
  year = 2023,
  month = jul,
  pages = {2575--2588},
  publisher = {Association for Computational Linguistics},
  address = {Toronto, Canada},
  doi = {10.18653/v1/2023.findings-acl.161},
  urldate = {2023-08-18}
}

@article{fleming2024MedAlignClinicianGeneratedDataset,
  title = {{{MedAlign}}: {{A Clinician-Generated Dataset}} for {{Instruction Following}} with {{Electronic Medical Records}}},
  shorttitle = {{{MedAlign}}},
  author = {Fleming, Scott L. and Lozano, Alejandro and Haberkorn, William J. and Jindal, Jenelle A. and Reis, Eduardo and Thapa, Rahul and Blankemeier, Louis and Genkins, Julian Z. and Steinberg, Ethan and Nayak, Ashwin and Patel, Birju and Chiang, Chia-Chun and Callahan, Alison and Huo, Zepeng and Gatidis, Sergios and Adams, Scott and Fayanju, Oluseyi and Shah, Shreya J. and Savage, Thomas and Goh, Ethan and Chaudhari, Akshay S. and Aghaeepour, Nima and Sharp, Christopher and Pfeffer, Michael A. and Liang, Percy and Chen, Jonathan H. and Morse, Keith E. and Brunskill, Emma P. and Fries, Jason A. and Shah, Nigam H.},
  year = 2024,
  month = mar,
  journal = {Proceedings of the AAAI Conference on Artificial Intelligence},
  volume = {38},
  number = {20},
  pages = {22021--22030},
  issn = {2374-3468},
  doi = {10.1609/aaai.v38i20.30205},
  urldate = {2024-10-18},
  copyright = {Copyright (c) 2024 Association for the Advancement of Artificial Intelligence},
  langid = {english}
}

\includepdf[pages=-, pagecommand={\thispagestyle{fancy}}]{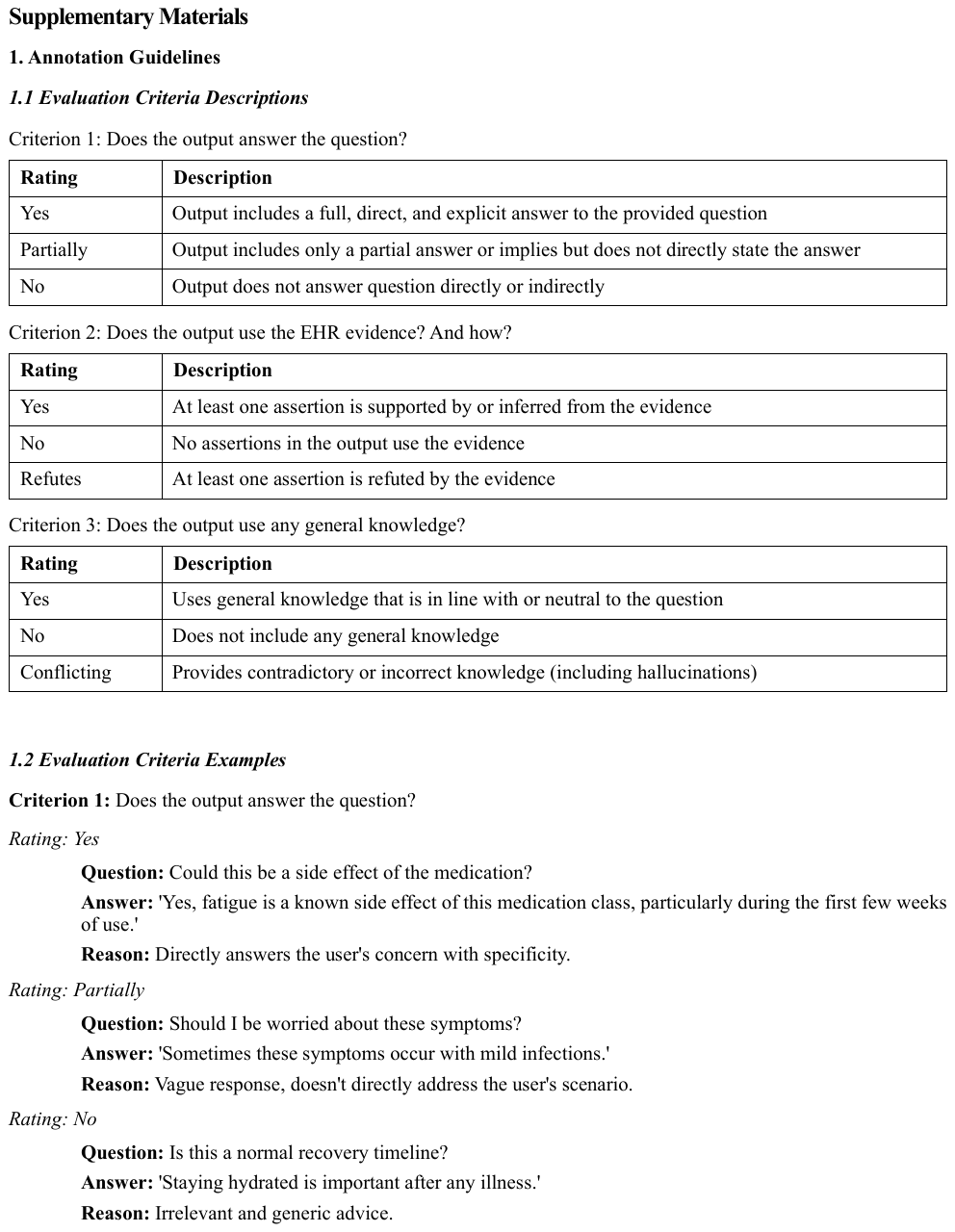}

\end{document}